\documentclass[review,12pt]{elsarticle}
\usepackage{lineno,hyperref}
\usepackage{array}
\usepackage{amssymb}
\usepackage{enumerate}
\usepackage{makecell}
\usepackage{mathtools}
\usepackage{array}
\usepackage{multirow}
\usepackage{ulem}
\usepackage{color}
\usepackage{diagbox}
\usepackage{enumerate}
\usepackage{graphicx}
\usepackage{subfigure}
 \usepackage{tikz,mathpazo} 
\usepackage{indentfirst}
\usepackage{soul}
\usepackage{amsmath}
\usepackage[justification=centering]{caption}
\usepackage{tabularx}
\usepackage{algorithm}
\usepackage{algorithmicx}
\usepackage{algpseudocode}
\usepackage{adjustbox}
\usepackage{geometry}
\usepackage{booktabs}
\usetikzlibrary{shapes.geometric, arrows}
\modulolinenumbers[5]

\soulregister{\cite}7
\soulregister{\ref}7
\biboptions{numbers,sort&compress}
\newcommand{\circledsmall}[1]{\lower.7ex\hbox{\tikz\draw (0pt, 0pt)%
    circle (.5em) node {\makebox[0.1em][c]{\small#1}};}}

\newcommand{\circledtiny}[1]{\lower.7ex\hbox{\tikz\draw (0pt, 0pt)%
    circle (.3em) node {\makebox[0.1em][c]{\tiny #1}};}}

\journal{Journal of \LaTeX\ Templates}

\begin{document}

\begin{frontmatter}

\title{A Novel Method for Pignistic Information Fusion \\in the View of Z-number}

\author[]{Yuanpeng He}
%    \author[label1]{Yong Deng\corref{cor1}}
%    
%   \affiliation[label1]{organization={Institute of Fundamental and Frontier Science, University of Electronic Science and Technology of China},
%   	city={Chengdu},
%   	postcode={610054},
%   	country={China}}
   	
%    \affiliation[label2]{organization={School of Computer Science, Peking University},
%    	city={Peking},
%    	postcode={100871},
%    	country={China}}

%    \cortext[cor1]{Corresponding author: Yong Deng,  Institute of Fundamental and Frontier Science, University of Electronic Science and Technology of China, 610054, Chengdu, China; School of Medicine, Vanderbilt University, Nashville,  37240, Tennessee, USA. Email address:  dengentropy@uestc.edu.cn and prof.deng@hotmail.com}

\begin{abstract}
How to properly fuse information from complex sources is still an open problem. Lots of methods have been put forward to provide a effective solution in fusing intricate information. Among them, Dempster-Shafer evidences theory (DSET) is one of the representatives, it is widely used to handle uncertain information. Based on DSET, a completely new method to fuse information from different sources based on pignistic transformation and Z-numbers is proposed in this paper which is able to handle separate situations of information and keeps high accuracy in producing rational and correct judgments on actual situations. Besides, in order to illustrate the superiority of the proposed method, some numerical examples and application are also provided to verify the validity and robustness of it.
\end{abstract}


\end{frontmatter}

\section{Introduction}
The world is inundated with lots of information and new information is
produced from time to time. As a result, how to appropriately combine information from complex sources has drawn a lot of attention. Researchers
around the world have put considerable efforts in extracting truly valuable part of information and relative theories have been proposed to satisfy the urgent need. The representatives of them are maximum theories \cite{DBLP:journals/ijccc/Gao020,DBLP:journals/tfs/Yager15a,DBLP:journals/tfs/Xiao21a}, fuzzy theory \cite{DBLP:journals/air/GargR20,DBLP:journals/ijccc/DengD20,he2024matrixbaseddistancepythagoreanfuzzy,DBLP:books/sp/Atanassov99,DBLP:journals/asc/SongFWW19,he2022ordinal}, complex mass function \cite{DBLP:journals/tcyb/Xiao22,DBLP:journals/ijis/GargR19,he2023tdqmf,DBLP:journals/apin/Xiao20,DBLP:journals/jifs/Xiao20}, soft theory \cite{DBLP:journals/ijis/FeiFL19,DBLP:journals/tfs/AlcantudFY20,DBLP:journals/kbs/FengCPFH16,DBLP:journals/ijis/SongD19}, Dempster-Shafer evidence theory and its extension \cite{DBLP:journals/isci/DengJ20,he2021conflicting,DBLP:journals/asc/LiuZ20,DBLP:journals/ijis/LuoD20,he2022mmget,DBLP:journals/tfs/LuoD20}, Z-numbers \cite{DBLP:journals/tfs/JiangCD20,DBLP:journals/eaai/LiuTK19,DBLP:journals/isci/HeD23,DBLP:journals/cam/LiPD20} and D numbers \cite{DBLP:journals/ijis/DengJ19,DBLP:journals/ijfs/Deng019,DBLP:journals/ijitdm/LiuZW20,DBLP:journals/jifs/Mo21}. For these theories, they all try to solve
the problem that how to obtain effective and clear judgments from uncertain and intricate information and different methods are offered based on
them. Because of the effectiveness of these theories, they are also be utilized in many applications, like pattern recognition \cite{he2024generalized,DBLP:journals/tfs/LiuLDC20,DBLP:journals/tsmc/Xiao21,he2022new}, risk evaluation \cite{DBLP:journals/kbs/FuXCXY20,DBLP:journals/inffus/PanZWS20,DBLP:journals/tfs/PanZLD20,DBLP:journals/tits/WangFZ21,DBLP:conf/ecsqaru/Soubaras09}, decision making \cite{DBLP:journals/tfs/XiaoCJ21,DBLP:journals/symmetry/GongF21,he2023ordinal,DBLP:journals/apin/HeX22,DBLP:journals/tcyb/TangLHCP21} and some other applications \cite{DBLP:journals/chinaf/Deng20,DBLP:journals/kbs/LiuWD20,DBLP:journals/ijis/ZhanX21}.
Among them, Dempster-Shafer evidence theory (DSET) is a powerful
tool in managing uncertain information and multi-source information fusion. Currently, the various methods for multi-sensor information fusion based on evidence theory are broadly divided into two categories. The first category mainly utilizes the entropy weight method to weigh the evidence to be fused. Specifically, in evidence theory, different pieces of evidence (or information sources) may contribute differently to the decision-making process \cite{DBLP:journals/isci/WuX24, DBLP:journals/eaai/LiuLD24, DBLP:journals/eaai/Deng023}. The entropy weight method evaluates the importance or reliability of each piece of evidence by calculating its entropy value. Evidence with lower entropy values (i.e., greater information variability) is considered more informative and thus is given greater weight in the fusion process. The entropy weight method provides an objective method of assigning weights to different pieces of evidence based on their information content \cite{DBLP:journals/isci/PanG23, DBLP:journals/soco/HeD23}. This is crucial in evidence theory, as subjective judgment can lead to bias. Furthermore, in cases of evidence conflict, the entropy weight method helps prioritize more likely reliable evidence, thereby reducing the impact of less reliable conflicting information. In summary, the role of the entropy weight method in information fusion in evidence theory lies in its systematic and objective approach to assess and integrate multiple pieces of evidence, especially under conditions characterized by uncertainty and incomplete information, thereby enhancing the effectiveness of the decision-making process.

The second category involves using different credibility distances to measure the similarity between pieces of evidence, followed by a fusion process based on similarity \cite{DBLP:journals/ijis/ZhangX22, DBLP:journals/fss/ZhouZYW24, DBLP:journals/apin/LuX24}. Evidence distance is used to measure and compare the differences and consistencies between different sources of evidence, where smaller distances indicate greater consistency between the evidence, and larger distances indicate greater inconsistency. In evidence theory, conflict between evidence is a very important research focus. During the fusion process, if the distance between two sources of evidence is large, it indicates that they may have significant conflict. By identifying and analyzing these conflicts, inconsistencies in information fusion can be more effectively addressed and resolved \cite{DBLP:journals/ijis/HeX21, DBLP:journals/cam/HeD22}. This can also improve the accuracy and reliability of information fusion. The fusion process can generate more comprehensive and credible conclusions based on considering the uniqueness and differences of each piece of evidence. For decision support systems, considering the distance between pieces of evidence can help decision-makers better understand the relationships and reliability of different information sources, thereby making wiser decisions. However, both methods are not without flaws. For example, the entropy weight method mainly considers the variability of each indicator when calculating weights, but it does not consider the correlation between data. This may lead to the neglect of some important information, especially when there is a high correlation between indicators. The method of using distance for evidence fusion can lead to unstable or inaccurate results when there is very high conflict between the evidence. In extreme cases, these conflicts can cause the fusion process to fail. 

To solve these problems, this paper proposes solutions from two aspects. We separate the conflicting parts in evidence fusion and calculate the corresponding uncertainty for each piece of evidence to determine the weight between the evidence, addressing potential extreme conflicts in evidence fusion. To further enhance the reliability of evidence fusion, we refer to the concept of Z-number, constructing a Z-number-like structure for each piece of evidence to improve overall credibility. Then, we use the pignistic transformation to resolve the uncertainty of the evidence and innovatively propose the use of OWA operators to fuse the constructed Z-numbers, obtaining the corresponding credibility value for each proposition. Finally, using the traditional Dempster combination rule, the processed evidence is fused to obtain the credibility assignment for each proposition. In sum, the method proposed in this paper takes advantage of DSET and could produces much more accurate and reasonable results by redesigning the process of combination of information. And the main contribution of the method can be listed as:
\begin{enumerate}
	\item A reliability measure system is introduced into the process of combination so as to make clear the level of priority of information sources.
	\item By utilizing a counter-standardized rule of obtaining conflicting part,
	a consistency in results between traditional Dempster combination rule and proposed method is achieved.
	\item A concise aggregation of information is satisfied by taking advantage of the concept of OWA which ensures completeness of information processed.
	\item A much better and clearer performance in generating final judgments can be realized which defeats traditional and some modern methods under different circumstances.
\end{enumerate}

And the rest of this paper is written as follows. In the section of preliminaries, some relative concepts are generally introduced. Besides, detailed process is illustrated in the part of methodology. Moreover, some
numerical examples and application are provided to prove the correctness
and validity of method offered in this paper in the section of numerical examples and application respectively. Conclusions are made in the part of conclusion.

\section{Preliminaries}
In this part, some related concepts are briefly introduced and some interesting works have been constructed by utilizing them \cite{DBLP:journals/tfs/TianLMK21,DBLP:journals/ijccc/PanD20,DBLP:journals/ijcisys/LiaoRF20,DBLP:journals/tfs/Yager19,DBLP:journals/access/CaiGD20}.
\subsection{Dempster-Shafer evidence theory \cite{DBLP:series/sfsc/Dempster08a,DBLP:journals/ijar/Shafer16}}
Assume set $Y$ is a non-empty set and finite set which contains N elements which are mutually exclusive. Therefore, the limited set is named as frame of discernment (FOD). And $Y$ is defined as:
\begin{equation}
	Y = \{V_1,V_2,V_3,\cdots,V_N\}
\end{equation}

Based on the definition of FOD, then the power set of it can be defined as:
\begin{equation}
	2^{Y} = \{V_1,V_2,\cdots,\{V_1,V_2\},\cdots,\{V_1,\cdots,V_N\}\}
\end{equation}

Then, a concept of basic probability assignment (BPA) can be designed on the definition introduced before which is denoted as $\xi$ and can be defined as:
\begin{equation}
    \xi: 2^{Y} \rightarrow [0,1]
\end{equation}

Besides, the properties of the function developed on the FOD can be defined as:
\begin{equation}
	\xi(\emptyset) = 0
\end{equation}
\begin{equation}
	\sum_{V \in 2^{Y}}\xi(V) = 1
\end{equation}

When proposition $V$ satisfies the properties proposed above, then the mass of $\xi(V)$ indicates the belief support to the proposition provided by evidences. If the corresponding mass of a certain proposition gets higher, then the credibility of the incident represented by the proposition becomes higher. Vice versa. Besides, when the mass is bigger than $0$, then it can be called as a focal element.

Moreover, it is very usual to encounter the situation that there exists different sources of evidences which are expected to be fused to get a final judgment. Because the source of the information is independent, Dempster adopts an orthogonal method to combine all of evidences which is commutative and is defined as:
\begin{equation}
	\xi(A) = \frac{1}{1-\varrho}*\sum_{Q_i \cap W_i \cap R_p \cap \cdots = A} \xi_1(Q_1) * \xi_2(W_l) * \xi_3(R_p) * \cdots
\end{equation}

And the coefficient of the formula, $\varrho$, is a representative of the degree of conflict of the evidences provided which can be defined as:
\begin{equation}
	\varrho = \sum_{Q_i \cap W_i \cap R_p \cap \cdots = \emptyset} \xi_1(Q_1) * \xi_2(W_l) * \xi_3(R_p) * \cdots
\end{equation}

\subsection{Deng Entropy \cite{DBLP:journals/chinaf/Deng20}}
Assume there exists a series of BPAs, Deng entropy can be defined as:
\begin{equation}
	E_D = - \sum_{R \in 2^{Y}}\xi(R)log_2 \frac{\xi(R)}{2^{|R|}-1}
\end{equation}

where the $\xi$ is a BPA defined on the FOD $Y$ and $R$ is a focal element of
$\xi$. Besides, $|\xi|$ indicates the the carnality of the proposition.

\subsection{Pignistic transformation \cite{DBLP:journals/ijar/Smets05}}
Assume there exists a mass function defined on the FOD $Y$, the corresponding pignistic probability transformation (PPT), $BetP_{\xi(V)} : Y \rightarrow [0, 1]$, can be defined as:
\begin{equation}
	BetP_{\xi(V)} = \sum_{R \subseteq Y, V \in R} \frac{1}{|R|}\frac{\xi(R)}{1 - \xi(\emptyset)}
\end{equation}

where the $|R|$ indicates the carnality of proposition $V$ and $\xi(\emptyset)$ is not equal to $0$. Besides, the mass function can be transformed into the form of probability distribution.

\subsection{Ordered weighted averaging aggregation \cite{DBLP:series/sfsc/Yager11}}
OWA operator is proposed to provide a method to optimize the effect of clustering. It can be also regarded as a mapping $A_n \rightarrow A$ which is associated with a vector with $n$ dimension and can be defined as:
\begin{equation}
	OWA=\left\{
	\begin{aligned}
		\vartheta_1 \\
		\cdots \\
	    \cdots	\\
		\vartheta_1
	\end{aligned}
	\right\}
\end{equation}
\begin{equation}
	OWA(q_1,q_2,\cdots,q_n) = \sum_{y=1}^{n}\vartheta_y w_y
\end{equation}

where $w_y$ is the $e$th largest value of $q_e$. And the property it satisfies can be defined as:
\begin{equation}
	\sum_{i=1}^{n} \vartheta_i; 0 \leq\vartheta\leq1
\end{equation}

If the $OWA = OWA^{*} = [1, 0, ..., 0]^{T}$, it is designed for an optimized strategy in decision making; If the $OWA = OWA^{*} = [0, 0, ..., 1]^{T}$, it is designed for a pessimistic one. In a unbiased situation, $OWA = OWA_U = [ \frac{1}{n} , \frac{1}{n}, ...,\frac{1}{n} ]^{T}.$

\subsection{Yager's rule of combination \cite{DBLP:journals/isci/Yager87a}}
In order to indicate the degree of conflict among evidences and avoid produce counter-intuitive judgments of current situations, Yager proposed a new rule of combination of evidences. The rule can be defined as:
\begin{equation}
	\xi(\emptyset) = 0
\end{equation}
\begin{equation}
	\xi(B) = \sum_{Z_a \cap X_s \cap C_d \cap \cdots = B} \xi_1(Z_a) * \xi_2(X_s) * \xi_3(C_d) * \cdots
\end{equation}
\begin{equation}
	\varrho = \sum_{Z_a \cap X_s \cap C_d \cap \cdots = \emptyset} \xi_1(Z_a) * \xi_2(X_s) * \xi_3(C_d) * \cdots
\end{equation}
\begin{equation}
	\xi(Con) = \sum_{Z_a \cap X_s \cap C_d \cap \cdots = Con} \xi_1(Z_a) * \xi_2(X_s) * \xi_3(C_d) * \cdots + \varrho
\end{equation}

It should be pointed out that the value of $\xi(Con)$ is simply utilized to indicate the degree of conflict and no use in describing actual situations. However, it can not avoid the phenomenon that counter-intuitive results are produced fundamentally.

\subsection{Z-number \cite{DBLP:journals/isci/Zadeh11}}
The Z-number is composed of two fuzzy numbers. With respect to a random variable $I$, the Z-number can be constructed as follows. The first part of the number is a direct judgment on the variable $I$. Besides, the other part of the number is an estimate on the first part which is also regarded as a kind of reliability. The mathematics form of Z-number can be defined as:
\begin{equation}
	Z = (C, D)
\end{equation}

The concept of Z-number is proposed by Zadeh which is utilized to model uncertain information. And the part of $C$ and $D$ are not completely mutually independent.

\section{Proposed method}
Detailed process of proposed method is provided in this section.
\subsection{Obtain conflicting part of evidences}
The Yager fusion rule is a method used for information fusion in evidence theory (Dempster-Shafer theory). Proposed by Ronald R. Yager, it aims to address issues that may arise with Dempster combination rule when dealing with highly conflicting evidence. By allocating the impact of conflicting evidence to a specific "uncertainty" category, the results of the Yager fusion rule are typically more intuitive and easier to understand. This assists users in better interpreting and comprehending the fusion outcomes. Compared to some other fusion methods, the Yager rule is able to preserve more of the original information during the fusion process, which can be very important for subsequent analysis and decision-making. Taking advantage of Yager’s rule of combination, a simplified rule to obtain conflicting part is designed. First, combine evidences utilizing traditional Dempster combination rule without a step of normalization to avoid to exaggerate underlying mistaken effects due to highly conflicting
phenomenon but remain consistent with traditional solution in combination as far as possible. Then, collect undistributed part of mass contained in BPA as the part indicating conflicts. The process can be defined as:
\begin{equation}
	\xi(P_i)^{C} = \sum_{I_r \cap O_t \cap P_y \cap \cdots = P_i} \xi_1(I_r) * \xi_2(O_t) * \xi_3(P_y) * \cdots
\end{equation}
\begin{equation}
	\xi(X) = 1 - \sum_{i=1}^{n}\xi(P_i)^{C}
\end{equation}

This operation remains a basic ability in indicating underlying target benefiting from Yager’s method. In the Dempster combination rule, there is a requirement for the normalization of probabilities, which can pose issues when faced with highly conflicting evidence. The approach proposed in this paper is akin to the Yager rule in its method of handling conflicts. It avoids the problems arising from such normalization through a pre-processing step before the fusion of evidence.
\subsection{Calculate degree of uncertainty of evidences}
Deng entropy, by considering specific characteristics of evidence, provides a more precise measure of uncertainty compared to traditional entropy methods. This results in a more in-depth and accurate analysis of the evidence. It is particularly well-suited for dealing with complex or incomplete information. Deng entropy can effectively measure uncertainty in these scenarios, aiding in a better understanding and handling of complex data. In sum, Deng entropy is an efficient tool in measuring uncertainty of FOD, so it is utilized to provide a precise description of current situations. Assume certain propositions are contained in $Evi_i$ and the process can be defined as:
\begin{equation}
	E_{D}^{Evi_i} = - \sum_{P_i \in 2^{Y}} \xi(P_i)log_2 \frac{\xi(P_i)}{2^{P_i} - 1}
\end{equation}

For every piece of evidence, the degree of uncertainty can be regarded as a believable judgment on reliability measure which also lays the foundation for the subsequent construction of corresponding Z-numbers based on the related concept.

\subsection{Generate Z-numbers for evidences utilizing Deng entropy}
Z-numbers, introduced by Lotfi A. Zadeh, enhance the representation and processing of uncertain information by incorporating both the degree of certainty and the reliability of that information. The dual-structure approach aligns closely with natural human expression and understanding of uncertainty, making Z-numbers particularly useful in decision-making scenarios where precise data may not be available or the information is subjective. Their adaptability across various domains, from engineering to social sciences, stems from their ability to model different types of uncertainties, extending the capabilities of fuzzy logic with an added layer of reliability information. This makes Z-numbers especially valuable in complex environments and risk assessment, where they provide a nuanced understanding of uncertainty, addressing the limitations of traditional numerical data in capturing the intricacies of real-world information. Besides, Deng entropy is widely used to measure degree of uncertainty, so the opposite of it is introduced into the credibility measure, which remains consistent with the concept of Z-number. Assume any piece of evidence is denoted as $Evi_i$ and the process of obtaining Z-numbers can be defined as:
\begin{equation}
	\Delta_{Evi_i} = \frac{e^{-E_{D}^{Evi_i}}}{\sum_{k=1}^{n}e^{-E^{Evi_k}}_{D}}
\end{equation}

Assume an expanded form a piece of an evidence can be given as $Evi_i = \{A_1,A_2,\cdots,\{A_1,\cdots,\\ A_N\}\}$. Then, corresponding Z-numbers can be constructed
as follows:

$Z_{Evi_1} = \{Z_{A_1} = (\xi(A_1), \Delta_{Evi_1}), Z_{A_2} = (\xi(A_2), \Delta_{Evi_1}),\cdots,Z_{\{A_1,\cdots,A_N\}} = (\xi(\{A_1,\cdots,A_N\}),\Delta_{Evi_1})\}$

$Z_{Evi_2} = \{Z_{A_1} = (\xi(A_1), \Delta_{Evi_2}), Z_{A_2} = (\xi(A_2), \Delta_{Evi_2}),\cdots,Z_{\{A_1,\cdots,A_N\}} = (\xi(\{A_1,\cdots,A_N\}),\Delta_{Evi_2})\}$

$\quad \vdots \quad \qquad  \qquad  \qquad  \qquad  \qquad  \qquad   \vdots \quad \qquad  \qquad  \qquad  \qquad \vdots \qquad  \qquad \vdots$

$Z_{Evi_n} = \{Z_{A_1} = (\xi(A_1), \Delta_{Evi_n}), Z_{A_2} = (\xi(A_2), \Delta_{Evi_n}),\cdots,Z_{\{A_1,\cdots,A_N\}} = (\xi(\{A_1,\cdots,A_N\}),\Delta_{Evi_n})\}$

Moreover, it is very necessary to take reliability of evidences into consideration to alleviate negative influences brought by some extremely abnormal evidences in order to produce much more intuitive and reasonable assessments. In this section, the evidences which possesses higher degree of uncertainty is regarded more unreliable. Overall, most existing work involves obtaining an uncertainty assessment for each piece of evidence, but we can refine this uncertainty assessment down to each individual proposition. Then, each proposition is first fused in the form of a Z-number, which will be explained in the following section.

\begin{figure*}[h]
	\centering
	\includegraphics[scale=1.2]{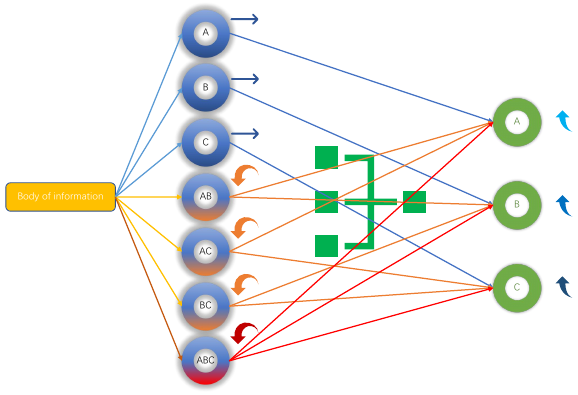}
	\caption{Process of evaluating hidden probability}
	\label{Evaluate}
\end{figure*}

\subsection{Evaluate hidden probability by simplified pignistic transformation}
The pignistic transformation in belief function theory is a powerful tool for decision-making under uncertainty, as it effectively converts belief degrees into a more usable probability distribution. This transformation bridges the gap between the theoretical aspects of Dempster-Shafer theory and practical decision-making by providing a pragmatic approach that balances the available evidence and the inherent risks of uncertain environments. It is particularly useful in mitigating conflicts in evidence sources, making it compatible with expected utility theory and facilitating a more stable and intuitive basis for decisions. By aggregating various pieces of evidence and reflecting a compromise in situations with partial or incomplete information, the pignistic transformation renders belief function theory more applicable and relevant in real-world decision-making scenarios. For any group of Z-numbers, if there exist multiple propositions then allocate the value of multiple propositions to single propositions to reduce degree of uncertainty and make clearer decisions. Assume certain proposition belongs to corresponding evidence $Evi_k$ and the process can be defined as:
\begin{equation}
	Z_{P_i} = (\sum_{U \subseteq Y, P_i \in U}\frac{\xi(U)}{|U|}, \Delta_{Evi_k}) = (\eta(P_i)^{Evi_k},\Delta_{Evi_k})
\end{equation}

By taking this step, the real information volume can be extracted and
be used for information management and processing. The necessity of carrying this kind of operation is thoroughly illustrated in \cite{DBLP:journals/ijar/Smets05}. For each singleton subset proposition, its corresponding uncertain information is allocated to itself, which aids in more definitive decision-making. Meanwhile, to ensure the reliability of each proposition, we construct a corresponding Z-number for each, to be used in the subsequent reliable fusion process and the detailed process is illustrated in Fig \ref{Evaluate}.

\subsection{Combine Z-numbers using concept of OWA}
In this section, all of the values corresponding to certain propositions in Z-numbers are combined and summed which are obtained to divide the mass of conflicting part proportionally. OWA is a flexible and powerful tool that allows for the aggregation of multiple inputs while considering their relative importance and the decision maker's attitude towards risk or uncertainty. Due to the fact that the Z-numbers corresponding to each proposition in the constructed evidence precisely fit the form of the Ordered Weighted Averaging (OWA) process, we propose a decision-making process that utilizes this approach to fuse different Z-numbers. Assume certain proposition belongs to corresponding evidence $Evi_k$ the process can be defined as:
\begin{equation}
	\omega_{P_i}=\{\Delta_{Evi_1}, \cdots, \Delta_{Evi_m} \cdots, \Delta_{Evi_n}\}
\end{equation}
\begin{equation}
	\varepsilon(P_i) = \sum_{k=1}^{n}\eta(P_i)^{Evi_k} \times \omega_{P_i}
\end{equation}

The process of combing Z-numbers also highlight the motivation of OWA, it distributes different weights to different propositions to select the ones which is more preferable and useful for information fusion. By adjusting the weights of the inputs, the OWA operator can reduce the impact of extreme values or outliers, thereby reducing the bias in the aggregated result.

\subsection{Obtain final values of propositions}
Using the results of combination and the values obtained in the process of getting conflicting part, the final values of propositions are assembled and the process can be defined as:
\begin{equation}
	\xi(P_i)_{Modified} = \varepsilon(P_i) \times \xi(X) + \xi(P_i)^{C}
\end{equation}

Then, the modified values of evidences of different propositions are supposed to by synthesized into one evidence which is a preparation for the next step. In essence, it entails combining the outcomes derived from the Z-number fusion with those segments of the original evidence that are conflict-free, as retained based on Yager's fusion rule principles.

\begin{figure*}[h]
	\centering
	\includegraphics[scale=1.2]{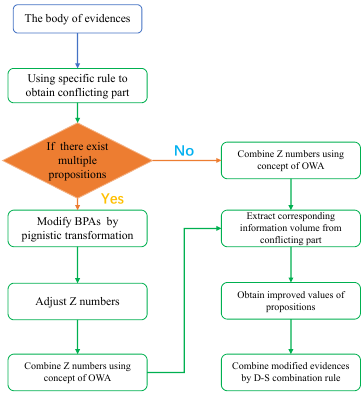}
	\caption{The detailed process of proposed method}
	\label{detailed}
\end{figure*}

\subsection{Combine modified evidences}
Dempster's rule of combination provides a powerful and flexible approach for aggregating evidence, handling uncertainty, and supporting decision-making processes in a wide range of applications. But it's also important to note that the rule has limitations, especially when dealing with highly conflicting evidence sources, where it might lead to counter-intuitive results. However, after the processing in the previous steps, the evidence information currently handled no longer has the potential for high conflict, and traditional fusion rules can effectively combine evidence from different sources. Then, combine the modified values of propositions using traditional Dempster combination rule $n-1$ times to get final judgment of actual situations. Moreover, the whole fusion process is provided in Fig \ref{detailed}.

\section{Numerical examples}
\subsection{\textbf{Example 1:} Extremely conflicting environment}
\begin{figure*}
	\centering
	\includegraphics[scale=1.2,angle=-90]{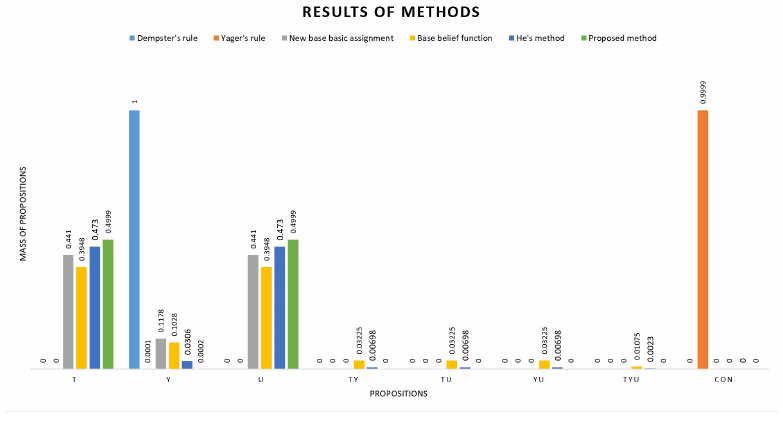}
	\caption{The detailed process of proposed method}
	\label{fig3}
\end{figure*}

Assume there exist two pieces of evidences and corresponding FOD is defined as $\Theta = \{T,Y, U\}$. And the detailed information about the two evidences is given in Table \ref{2222}. This example describes a phenomenon that if evidence highly conflicts with each other, traditional Dempster
combination rule may produce counter-intuitive results which is observed by Zadeh \cite{DBLP:journals/aim/Zadeh86}. Besides, the results of combination of six methods including proposed method are given in Table \ref{fig3-t} and Fig \ref{fig3}.

\begin{table}
	\centering
	\begin{tabular}{cccc}% 其中，tabular是表格内容的环境；c表示centering，即文本格式居中；c的个数代表列的个数
		%\toprule %[2pt]设置线宽     
		%a & b  &  c \\ %换行
		%\midrule %[2pt]  
		\bottomrule
		$Evidence_1$ & $\xi(T)$ & $\xi(Y)$ & $\xi(U)$  \\
		 & 0.99 & 0.01 & 0.00\\
		 $Evidence_2$& $\xi(T)$ & $\xi(Y)$ & $\xi(U)$ \\
		 &0.00 & 0.01 & 0.99\\
		\bottomrule %[2pt]     
	\end{tabular}
	\caption{Detailed information of two evidences in \textbf{Example 1}}
	\label{2222}
\end{table}

\begin{table}\scriptsize
	\centering
	\begin{tabular}{ccccccccc}% 其中，tabular是表格内容的环境；c表示centering，即文本格式居中；c的个数代表列的个数
		%\toprule %[2pt]设置线宽     
		%a & b  &  c \\ %换行
		%\midrule %[2pt]  
		\bottomrule
		$Dempster's\ rule$ \cite{DBLP:series/sfsc/Dempster08a} & $\xi(T)$ & $\xi(Y)$ & $\xi(U)$& $\xi(T,Y)$ & $\xi(T,U)$ &$\xi(Y,U)$& $\xi(T,Y,U)$ & $\xi(Con)$   \\
		& 0.00 & 1.00 & 0.00 & 0.00 & 0.00 & 0.00 & 0.00 & 0.00\\
		$Yager's\ rule$ \cite{DBLP:journals/isci/Yager87a} & $\xi(T)$ & $\xi(Y)$ & $\xi(U)$& $\xi(T,Y)$ & $\xi(T,U)$ &$\xi(Y,U)$& $\xi(T,Y,U)$ & $\xi(Con)$   \\
		& 0.00 & 0.01 & 0.00 & 0.00 & 0.00 & 0.00 & 0.00 & 0.9999\\
		$New\ base\ basic\ assignment$ \cite{DBLP:journals/apin/JingT21} & $\xi(T)$ & $\xi(Y)$ & $\xi(U)$& $\xi(T,Y)$ & $\xi(T,U)$&$\xi(Y,U)$ & $\xi(T,Y,U)$ & $\xi(Con)$   \\
		& 0.4410 & 0.1178 & 0.4410 & 0.00 & 0.00 & 0.00 & 0.00 & 0.00\\
		$Base\ belief\ function$ \cite{DBLP:journals/jaihc/WangZD19} & $\xi(T)$ & $\xi(Y)$ & $\xi(U)$& $\xi(T,Y)$ & $\xi(T,U)$ &$\xi(Y,U)$& $\xi(T,Y,U)$ & $\xi(Con)$   \\
		& 0.3948 & 0.1028 & 0.3948 & 0.03225 & 0.03225 & 0.03225 & 0.01075 & 0.00\\
		$He's\ method$ \cite{DBLP:journals/ijis/HeX21} & $\xi(T)$ & $\xi(Y)$ & $\xi(U)$& $\xi(T,Y)$ & $\xi(T,U)$ &$\xi(Y,U)$& $\xi(T,Y,U)$ & $\xi(Con)$   \\
		& 0.4730 & 0.0306 & 0.4730 & 0.00698 & 0.00698 & 0.00698 & 0.0023 & 0.00\\
		$Proposed\ method$ & $\xi(T)$ & $\xi(Y)$ & $\xi(U)$& $\xi(T,Y)$ & $\xi(T,U)$ &$\xi(Y,U)$& $\xi(T,Y,U)$ & $\xi(Con)$   \\
		& \textbf{0.4999} & 0.0002 & \textbf{0.4999} & 0.00 & 0.00 & 0.00 & 0.00 & 0.00\\
		\bottomrule %[2pt]     
	\end{tabular}
	\caption{Results of combination of methods in \textbf{Example 1}}
	\label{fig3-t}
\end{table}

It can be easily concluded that traditional Dempster combination rule produces counter-intuitive results of judgment. No matter $Evidence_1$ or $Evidence_2$, neither of them support the proposition $Y$. On the contrary, it is believed that propositions $T$ and $U$ are supposed to be distributed a mass close to 0.5 to indicate the probability it may take place. Therefore, the results of Dempster combination rule are unbelievable. Similar situations occur in Yager’s method. Although the method avoid distributing abnormal mass to proposition $Y$, it still fails to allocate reasonable BPA to propositions $T$ and $U$. As a result, the two traditional rules are not robust enough when disposing highly conflicting conditions. To address the problem mentioned above, some methods are designed and the representatives are New base basic assignment [54], Base belief function [55] and He's method [56]. All of them provide a reasonable solution in solving conflicts among evidences. However, with respect to this example, like what is mentioned before, the propositions $T$ and $U$ are supposed to be allocated a mass close to 0.5, because the two evidences provided almost completely support proposition $T$and $Y$ respectively. Compared with other methods, the proposed method successfully allocates reasonable mass 0.4999 to proposition $T$ and $U$, which conforms to actual situations and is far better than other methods in distributing mass to conflicting part of evidences. Two evidences slightly support proposition $Y$, but the mass of the proposition is too small and can be nearly ignored to make sure the judgment is intuitive and clear. All in all, the proposed method makes up the shortcoming existing in classic methods and performs much better than other modern methods in describing situations of combination under highly conflicting conditions.

\begin{figure*}
	\centering
	\includegraphics[scale=1.2, angle=-90]{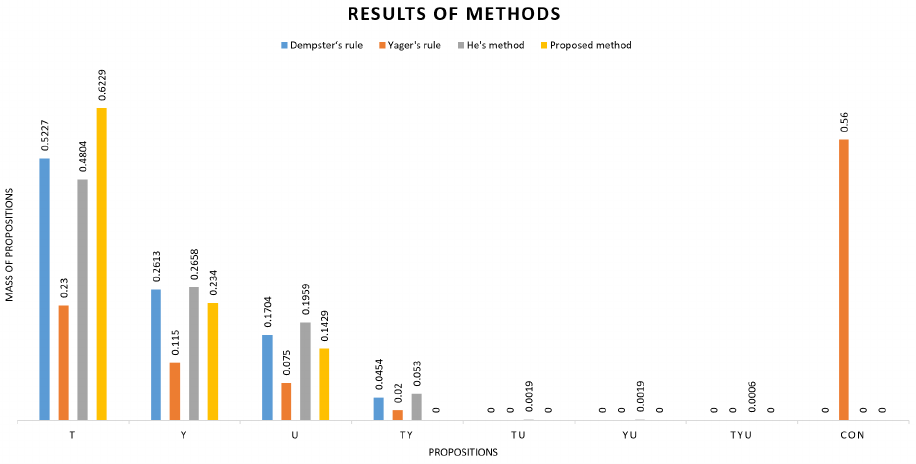}
	\caption{The detailed process of proposed method}
	\label{fig4}
\end{figure*}

\subsection{\textbf{Example 2}: Mild environment}
Assume there exist two pieces of evidences and corresponding FOD is defined as $\Theta = \{T,Y, U, \{T, Y\}\}$. And the detailed information about the two evidences is given in Table \ref{1111}. Besides, the results of combination of four methods including proposed method are given in Table \ref{fig4-t} and Fig \ref{fig4}.

\begin{table}
	\centering
	\begin{tabular}{ccccc}% 其中，tabular是表格内容的环境；c表示centering，即文本格式居中；c的个数代表列的个数
		%\toprule %[2pt]设置线宽     
		%a & b  &  c \\ %换行
		%\midrule %[2pt]  
		\bottomrule
		$Evidence_1$ & $\xi(T)$ & $\xi(Y)$ & $\xi(U)$ &  $\xi(T,Y)$ \\
		& 0.4 & 0.2 & 0.3 & 0.1\\
		$Evidence_2$& $\xi(T)$ & $\xi(Y)$ & $\xi(U)$ &  $\xi(T,Y)$ \\
		&0.3 & 0.25  & 0.25 & 0.2 \\
		\bottomrule %[2pt]     
	\end{tabular}
	\caption{Detailed information of two evidences in \textbf{Example 2}}
	\label{1111}
\end{table}

\begin{table}\scriptsize
	\centering
	\begin{tabular}{ccccccccc}% 其中，tabular是表格内容的环境；c表示centering，即文本格式居中；c的个数代表列的个数
		%\toprule %[2pt]设置线宽     
		%a & b  &  c \\ %换行
		%\midrule %[2pt]  
		\bottomrule
		$Dempster's\ rule$ \cite{DBLP:series/sfsc/Dempster08a} & $\xi(T)$ & $\xi(Y)$ & $\xi(U)$& $\xi(T,Y)$ & $\xi(T,U)$ &$\xi(Y,U)$& $\xi(T,Y,U)$ & $\xi(Con)$   \\
		& 0.5227 & 0.2613 & 0.1704 & 0.0454 & 0.00 & 0.00 & 0.00 & 0.00\\
		$Yager's\ rule$ \cite{DBLP:journals/isci/Yager87a} & $\xi(T)$ & $\xi(Y)$ & $\xi(U)$& $\xi(T,Y)$ & $\xi(T,U)$ &$\xi(Y,U)$& $\xi(T,Y,U)$ & $\xi(Con)$   \\
		& 0.23 & 0.1150 & 0.075 & 0.02 & 0.00 & 0.00 & 0.00 & 0.5600\\
		$He's\ method$ \cite{DBLP:journals/ijis/HeX21} & $\xi(T)$ & $\xi(Y)$ & $\xi(U)$& $\xi(T,Y)$ & $\xi(T,U)$ &$\xi(Y,U)$& $\xi(T,Y,U)$ & $\xi(Con)$   \\
		& 0.4804 & 0.2658 & 0.1959 & 0.0530 & 0.0019 & 0.0019 & 0.0006 & 0.00\\
		$Proposed\ method$ & $\xi(T)$ & $\xi(Y)$ & $\xi(U)$& $\xi(T,Y)$ & $\xi(T,U)$ &$\xi(Y,U)$& $\xi(T,Y,U)$ & $\xi(Con)$   \\
		& \textbf{0.6229} & 0.2340 & 0.1429 & 0.00 & 0.00 & 0.00 & 0.00 & 0.00\\
		\bottomrule %[2pt]     
	\end{tabular}
	\caption{Results of combination of methods in \textbf{Example 2}}
	\label{fig4-t}
\end{table}

It can be easily concluded that there exist no obvious conflicts among evidences in Example 2, so traditional Dempster combination rule produces intuitive results. However, the differences of mass of propositions are not distinct enough to produce straightforward and instant decision. What should be pointed out is that the results of traditional Dempster combination rule are generally acceptable and it is necessary to remain consistent with the trend of results of combination produced by it. Besides, Yager’s method indicates the degree of conflicts among evidences, but the mass of different propositions is not persuasive to be regarded as a proof
for decision making. And with respect to He’s method, it is obviously decreases the mass of proposition $T$ and less proper to make decisions, but like Yager’s method, the relationship of mass of propositions is not convincing compared with Dempster combination rule which is an obvious drawback and may leads to counter-straightforward results on judgment of practical situations. For proposed method, it improves the ability to distinguish valuable objects in decision making and remain consistent with distributions of mass of propositions. A mass of 0.6229 is allocated to proposition $T$ which is very easy for decision makers to have a choice. And the relationship of mass of proposition is the same as classic Dempster combination rule. In a word, the proposed not only improves the performance in indicating the proposition which is most possible to happen and do not destroy basic relationship of propositions, it can be regarded rational and validated therefore.

\begin{figure}
	\centering
	\includegraphics[scale=1.2, angle=-90]{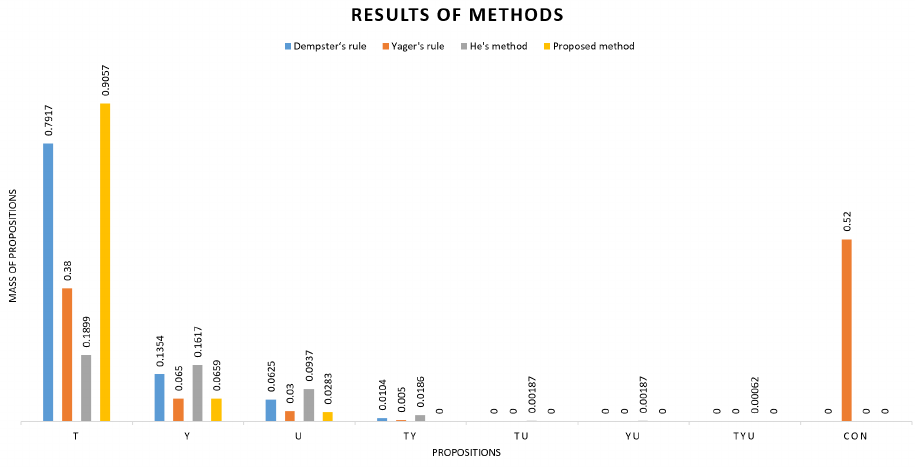}
	\caption{The detailed process of proposed method}
	\label{fig5}
\end{figure}

\subsection{\textbf{Example 3}: Mild environment}
Assume there exist two pieces of evidences and corresponding FOD is defined as $\Theta = \{T,Y, U, \{T,Y\}\}$. And the detailed information about the two evidences is given in Table \ref{dddd}. Besides, the results of combination of four methods including proposed method are given in Table \ref{fig5-t} and Fig \ref{fig5}.

\begin{table}
	\centering
	\begin{tabular}{ccccc}% 其中，tabular是表格内容的环境；c表示centering，即文本格式居中；c的个数代表列的个数
		%\toprule %[2pt]设置线宽     
		%a & b  &  c \\ %换行
		%\midrule %[2pt]  
		\bottomrule
		$Evidence_1$ & $\xi(T)$ & $\xi(Y)$ & $\xi(U)$ &  $\xi(T,Y)$ \\
		& 0.6 & 0.1 & 0.2 & 0.1\\
		$Evidence_2$& $\xi(T)$ & $\xi(Y)$ & $\xi(U)$ &  $\xi(T,Y)$ \\
		&0.5 & 0.3  & 0.15 & 0.05 \\
		\bottomrule %[2pt]     
	\end{tabular}
	\caption{Detailed information of two evidences in \textbf{Example 3}}
	\label{dddd}
\end{table}

\begin{table}\scriptsize
	\centering
	\begin{tabular}{ccccccccc}% 其中，tabular是表格内容的环境；c表示centering，即文本格式居中；c的个数代表列的个数
		%\toprule %[2pt]设置线宽     
		%a & b  &  c \\ %换行
		%\midrule %[2pt]  
		\bottomrule
		$Dempster's\ rule$ \cite{DBLP:series/sfsc/Dempster08a} & $\xi(T)$ & $\xi(Y)$ & $\xi(U)$& $\xi(T,Y)$ & $\xi(T,U)$ &$\xi(Y,U)$& $\xi(T,Y,U)$ & $\xi(Con)$   \\
		& 0.7917 & 0.1354 & 0.0625 & 0.0104 & 0.00 & 0.00 & 0.00 & 0.00\\
		$Yager's\ rule$ \cite{DBLP:journals/isci/Yager87a} & $\xi(T)$ & $\xi(Y)$ & $\xi(U)$& $\xi(T,Y)$ & $\xi(T,U)$ &$\xi(Y,U)$& $\xi(T,Y,U)$ & $\xi(Con)$   \\
		& 0.3800 & 0.0650 & 0.0300 & 0.0050 & 0.00 & 0.00 & 0.00 & 0.5200\\
		$He's\ method$ \cite{DBLP:journals/ijis/HeX21}& $\xi(T)$ & $\xi(Y)$ & $\xi(U)$& $\xi(T,Y)$ & $\xi(T,U)$ &$\xi(Y,U)$& $\xi(T,Y,U)$ & $\xi(Con)$   \\
		& 0.7215 & 0.1617 & 0.0937 & 0.0186 & 0.00187 & 0.00187 & 0.00062 & 0.00\\
		$Proposed\ method$ & $\xi(T)$ & $\xi(Y)$ & $\xi(U)$& $\xi(T,Y)$ & $\xi(T,U)$ &$\xi(Y,U)$& $\xi(T,Y,U)$ & $\xi(Con)$   \\
		& \textbf{0.9057} & 0.0659 & 0.0283 & 0.00 & 0.00 & 0.00 & 0.00 & 0.00\\
		\bottomrule %[2pt]     
	\end{tabular}
	\caption{Results of combination of methods in \textbf{Example 3}}
	\label{fig5-t}
\end{table}

The situation of this example is very similar to \textbf{Example 2}. In \textbf{Example 3}, all of relationship of the mass of propositions generated by methods are the same. The classic Dempster combination rule and Yager’s rule are not appropriate enough for decision making. Besides, He’s method is able to handle conflicting situations, but it produces even poorer result compared with Dempster combination rule, which is unacceptable under this circumstances. Moreover, proposed method addressed these problems well, it significantly improves the performance in indicating and the relationships of mass of propositions remain the same as traditional Dempster combination rule of combination. In one word, the method proposed in this paper effectively improves effect in indicating whether certain proposition is expected to take place and reduce degree of uncertainty of FOD to some extent to make it more straightforward.

\subsection{\textbf{Example 4: Mild environment}}
\begin{figure*}
	\centering
	\includegraphics[scale=1.2, angle=-90]{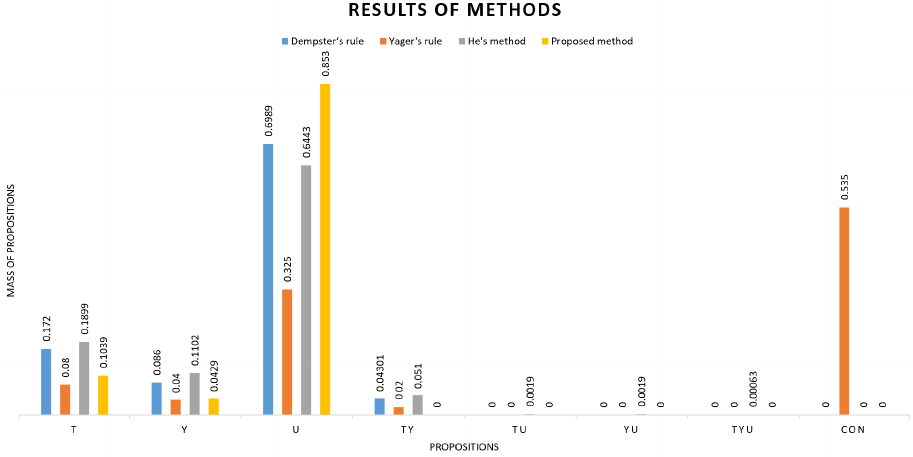}
	\caption{The detailed process of proposed method}
	\label{fig6}
\end{figure*}

Assume there exist two pieces of evidences and corresponding FOD is defined as $\Theta = \{T,Y, U, \{T,Y\}\}$. And the detailed information about the two evidences is given in Table \ref{22222}. Besides, the results of combination of four methods including proposed method are given in Table \ref{fig6-t} and Fig \ref{fig6}.

\begin{table}
	\centering
	\begin{tabular}{ccccc}% 其中，tabular是表格内容的环境；c表示centering，即文本格式居中；c的个数代表列的个数
		%\toprule %[2pt]设置线宽     
		%a & b  &  c \\ %换行
		%\midrule %[2pt]  
		\bottomrule
		$Evidence_1$ & $\xi(T)$ & $\xi(Y)$ & $\xi(U)$ &  $\xi(T,Y)$ \\
		& 0.2 & 0.1 & 0.5 & 0.2\\
		$Evidence_2$& $\xi(T)$ & $\xi(Y)$ & $\xi(U)$ &  $\xi(T,Y)$ \\
		&0.15 & 0.1  & 0.65 & 0.1 \\
		\bottomrule %[2pt]     
	\end{tabular}
	\caption{Detailed information of two evidences in \textbf{Example 4}}
	\label{22222}
\end{table}

\begin{table}\scriptsize
	\centering
	\begin{tabular}{ccccccccc}% 其中，tabular是表格内容的环境；c表示centering，即文本格式居中；c的个数代表列的个数
		%\toprule %[2pt]设置线宽     
		%a & b  &  c \\ %换行
		%\midrule %[2pt]  
		\bottomrule
		$Dempster's\ rule$ \cite{DBLP:series/sfsc/Dempster08a} & $\xi(T)$ & $\xi(Y)$ & $\xi(U)$& $\xi(T,Y)$ & $\xi(T,U)$ &$\xi(Y,U)$& $\xi(T,Y,U)$ & $\xi(Con)$   \\
		& 0.1720 & 0.0860 & 0.6989 & 0.04301 & 0.00 & 0.00 & 0.00 & 0.00\\
		$Yager's\ rule$ \cite{DBLP:journals/isci/Yager87a} & $\xi(T)$ & $\xi(Y)$ & $\xi(U)$& $\xi(T,Y)$ & $\xi(T,U)$ &$\xi(Y,U)$& $\xi(T,Y,U)$ & $\xi(Con)$   \\
		& 0.0800 & 0.0400 & 0.3250 & 0.02 & 0.00 & 0.00 & 0.00 & 0.5350\\
		$He's\ method$ \cite{DBLP:journals/ijis/HeX21}& $\xi(T)$ & $\xi(Y)$ & $\xi(U)$& $\xi(T,Y)$ & $\xi(T,U)$ &$\xi(Y,U)$& $\xi(T,Y,U)$ & $\xi(Con)$   \\
		& 0.1899 & 0.1102 & 0.6443 & 0.0510 & 0.00190 & 0.00190 & 0.00063 & 0.00\\
		$Proposed\ method$ & $\xi(T)$ & $\xi(Y)$ & $\xi(U)$& $\xi(T,Y)$ & $\xi(T,U)$ &$\xi(Y,U)$& $\xi(T,Y,U)$ & $\xi(Con)$   \\
		& 0.1039 & 0.0429 & \textbf{0.8530} & 0.00 & 0.00 & 0.00 & 0.00 & 0.00\\
		\bottomrule %[2pt]     
	\end{tabular}
	\caption{Results of combination of methods in \textbf{Example 4}}
	\label{fig6-t}
\end{table}

In this example, the proposed method still keeps highly accurate and concise standard in describing situations of FOD and is a very effective tool for decision makers to have choices in target recognition. All in all, the four examples have well illustrated that the proposed method not only can handle conflicting evidences felicitously and produce intuitive results, but also remain much better performance than traditional and other method in generating judgments with respect to evidences with normal relationships.

\section{Application}
All of the examples strongly support the credibility and effectiveness of proposed method. In order to illustrate the superiority of proposed method at a further step, an application applied in real data set under actual situations is designed. And the data set, iris, is drawn from UCI machine
learning repository.

\subsection{Application on IRIS dataset}
\begin{figure*}
	\centering
	\includegraphics[scale=1.2, angle=-90]{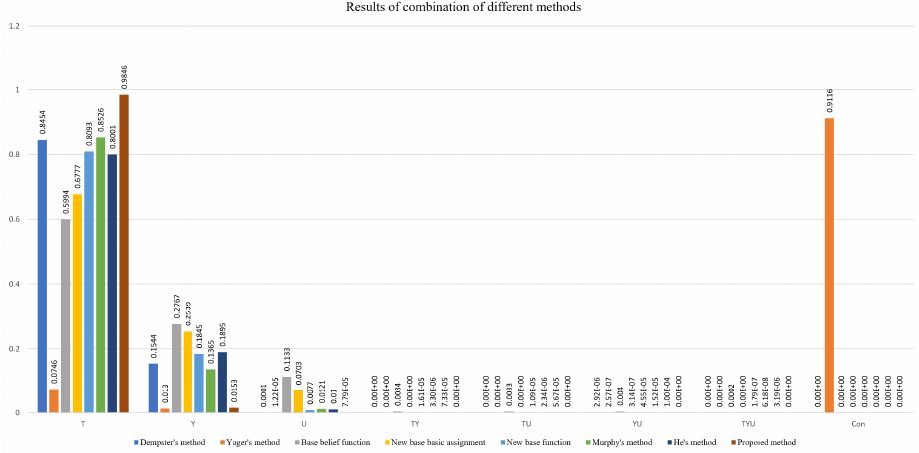}
	\caption{The detailed process of proposed method}
	\label{fig7}
\end{figure*}

Iris data set contains three categories which are (Setosa(T), Versicolor(Y), Virginica(U)) respectively. Besides, each categories possesses 50 instances. And 40 instances are chosen randomly to construct fuzzy triangles. And the remaining 10 instances are selected to generate BPAs. And the generated BPAs are provided in Table \ref{fdfdf}. Moreover, the results of combination of
different method are given in Table \ref{vvvv} and visualized results are in Fig \ref{fig7}.

\begin{table}\scriptsize
	\centering
	\begin{tabular}{cccccccc}% 其中，tabular是表格内容的环境；c表示centering，即文本格式居中；c的个数代表列的个数
		%\toprule %[2pt]设置线宽     
		%a & b  &  c \\ %换行
		%\midrule %[2pt]  
		\bottomrule
		$Attribute$& $\xi(T)$ & $\xi(Y)$ & $\xi(U)$& $\xi(T,Y)$ & $\xi(T,U)$ &$\xi(Y,U)$& $\xi(T,Y,U)$   \\
		$Sepal\ length$& 0.3337 & 0.3165 & 0.2816 & 0.0307 & 0.0052 & 0.0272 & 0.0052 \\
		$Sepal\ width$& 0.3164 & 0.2501 & 0.2732 & 0.0304 & 0.0481 & 0.0515 & 0.0304 \\
		$Petal\ length$& 0.6699 & 0.3258 & 0.00 & 0.00 & 0.00 & 0.0043 & 0.00\\
		$Petal\ width$& 0.6998 & 0.2778 & 0.00 & 0.00 & 0.00 & 0.0226 & 0.00 \\
		\bottomrule %[2pt]     
	\end{tabular}
	\caption{BPAs generated by utilizing IRIS dataset}
	\label{fdfdf}
\end{table}

\begin{table}\scriptsize
	\centering
	\begin{tabular}{ccccccccc}% 其中，tabular是表格内容的环境；c表示centering，即文本格式居中；c的个数代表列的个数
		%\toprule %[2pt]设置线宽     
		%a & b  &  c \\ %换行
		%\midrule %[2pt]  
		\bottomrule
		 $Method$& $\xi(T)$ & $\xi(Y)$ & $\xi(U)$& $\xi(T,Y)$ & $\xi(T,U)$ &$\xi(Y,U)$& $\xi(T,Y,U)$ & $\xi(Con)$   \\
		$Dempster's\ rule$ \cite{DBLP:series/sfsc/Dempster08a}& 0.8454 & 0.1544 & 0.0001 & 0.00 & 0.00 &  2.9197e-6  & 0.00 & 0.00\\
		$Yager's rule$ \cite{DBLP:journals/isci/Yager87a} & 0.0746 & 0.013 & 1.2243e-05 & 0.00 & 0.00 & 2.5676e-07 & 0.00 & 0.9116\\
		$Base\ belief\ function$ \cite{DBLP:journals/jaihc/WangZD19}& 0.5994 & 0.2767 & 0.1133 & 0.0034 & 0.0033 & 0.0040 & 0.0002 & 0.00\\
		$New\ base\ basic\ assignment$ \cite{DBLP:journals/apin/JingT21}& 0.6777 & 0.2539 & 0.0703 & 0.00 & 0.00 & 3.1395e-07 & 0.00 & 0.00\\
		$New\ base\ function$ \cite{DBLP:journals/apin/HeX22}& 0.8039 & 0.1845 & 0.0077 &  1.6074e-05&  1.0925e-05 &  4.5452e-05 & 1.7862e-07 & 0.00\\
		$Murphy's\ method$ \cite{DBLP:journals/dss/Murphy00} & 0.8526 & 0.1365 & 0.0121 & 3.3029e-06 & 2.3417e-06 & 1.5234e-05 & 6.1808e-08 & 0.00\\
		$He's\ method$ \cite{DBLP:journals/ijis/HeX21} & 0.8001 & 0.1895 & 0.0100 & 7.3275e-05 & 5.6740e-05 & 0.0001 & 3.1889e-06 & 0.00\\
		$Proposed\ method$& \textbf{0.9846} & 0.0153 & 7.7935e-05 & 0.00 & 0.00 & 0.00 & 0.00 & 0.00\\
		\bottomrule %[2pt]     
	\end{tabular}
	\caption{Results of combination of different methods}
	\label{vvvv}
\end{table}

\subsubsection{Discussions on results of combination of proposed method}
It can be concluded that the results of different methods indicate that proposition $T$ is the most possible to take place. And for other propositions, they are allocated much smaller mass. Therefore, in this case, the most urgent thing is to distinguish propositions $T$ from other propositions for decision makers. With respect to other propositions, like proposition $Y$, it is enough to distribute a relatively big mass to it compared with other propositions to manifest that proposition $Y$ is a second choice after selecting proposition $T$ as the most concerned object. And with respect to proposition $U$, there exist two evidences which hold a opinion that proposition $U$ is not supposed to happen. In multi-source information fusion, the condition is expected to be generally accepted to illustrate that proposition $U$ should be put into a very unnoticed position. And all of the concerns are well satisfied in the proposed method by allocating a mass of 0.9846 to proposition $T$ to present its overwhelming role in the information provided. Besides, propositions $Y$ and $U$ are also given reasonable mass, which conforms to basic judgments narrated before. All in all, the results combined by proposed method completely remain consistent with the ones generate by other methods and significantly improves the performance in indicating valuable part of complex information. As a result, it can be regarded as a powerful tool in managing information under intricate environment and offers a satisfying solution in multi-source information fusion.

\subsection{Application on target recognition}
In this section, the proposed method is utilized to solve problems about recognition of fault in machines and the corresponding information is extracted from \cite{DBLP:journals/prl/FanZ06}.

\subsubsection{Problem introduction}
The FOD for the categories of faults of machines is defined as $\chi = \{FA1, FA2, FA3\}$ and the group of sensors is also given as $S : \{Sen1, Sen2, Sen3\}$. Moreover, the information collected is modeled as BPAs and given in Table \ref{frfr} respectively.

\begin{table}\scriptsize
	\centering
	\begin{tabular}{ccccc}% 其中，tabular是表格内容的环境；c表示centering，即文本格式居中；c的个数代表列的个数
		%\toprule %[2pt]设置线宽     
		%a & b  &  c \\ %换行
		%\midrule %[2pt]  
		\bottomrule
		$Sensors$ & $\varsigma(FA_1)$ & $\varsigma(FA_1)$ & $\varsigma(Fa_2,FA_3)$& $\varsigma(FA_1,FA_2,FA_3)$    \\
		$Sen_1:\chi_1(\cdot)$& 0.60 & 0.10 & 0.10 & 0.20  \\
		$Sen_2:\chi_2(\cdot)$& 0.05 & 0.80 & 0.05 & 0.10  \\
		$Sen_3:\chi_3(\cdot)$& 0.70 & 0.10 & 0.10 & 0.10  \\
		\bottomrule %[2pt]     
	\end{tabular}
	\caption{The BPAs for fault information from multi sensors}
	\label{frfr}
\end{table}

\begin{table}\scriptsize
	\centering
	\begin{tabular}{ccccccc}% 其中，tabular是表格内容的环境；c表示centering，即文本格式居中；c的个数代表列的个数
		%\toprule %[2pt]设置线宽     
		%a & b  &  c \\ %换行
		%\midrule %[2pt]  
		\bottomrule
		$Method$ & $\varsigma(FA_1)$ & $\varsigma(FA_2)$ & $\varsigma(FA_3)$ & $\varsigma(FA_2,FA_3)$& $\varsigma(FA_1,FA_2,FA_3)$  &$Target$  \\
		$Dempster$ \cite{DBLP:series/sfsc/Dempster08a}& 0.4519 & 0.5048 & 0.00 & 0.0336 & 0.0096 &  $FA_2$  \\
		$Murphy's\ method$ \cite{DBLP:journals/isci/Yager87a}& 0.5410 & 0.4309 & 0.00 & 0.0215 &  0.0065 &  $FA_1$  \\
		$Fan\ and\ Zuo's\ method$ \cite{DBLP:journals/prl/FanZ06}&0.8119 & 0.1096 & 0.00 & 0.0526 & 0.0259 &  $FA_1$  \\
		$Yuan\ et\ al.$ \cite{DBLP:journals/sensors/YuanXFKD16}&  0.8948 & 0.0739 & 0.00 &  0.0241 & 0.0072 &  $FA_1$  \\
		$Xiao$ \cite{DBLP:journals/inffus/Xiao19}& 0.8973&  0.0688 & 0.00 & 0.0254 & 0.0080 &  $FA_1$  \\
		$Proposed\ method$& \textbf{0.9265} &  0.0720 & 0.0013 & 0.00 & 0.00 &  $FA_1$  \\
		\bottomrule %[2pt]     
	\end{tabular}
	\caption{Results of combination of different methods}
	\label{121212}
\end{table}

\begin{figure*}
	\centering
	\includegraphics[scale=1.2, angle=-90]{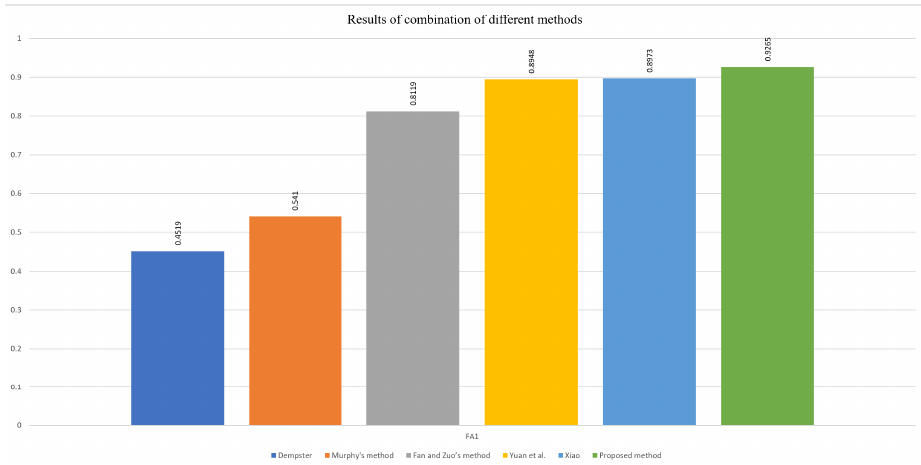}
	\caption{The detailed process of proposed method}
	\label{fig8}
\end{figure*}

\begin{figure*}
	\centering
	\includegraphics[scale=1.2, angle=-90]{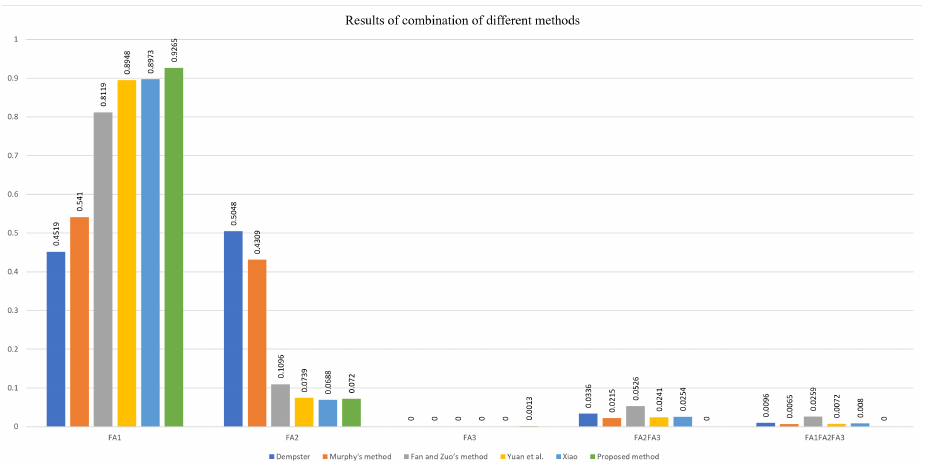}
	\caption{The detailed process of proposed method}
	\label{fig9}
\end{figure*}

\subsubsection{Some discussions on the results of combination of proposed method}
The combination results of different methods and corresponding visualizations are provided in Table \ref{121212}, Fig \ref{fig8} and \ref{fig9}.
It can be concluded that the traditional Dempster combination rule produces counter-intuitive results which recognize FA2 as the target while the rest of methods recognize FA1. Moreover, the proposed method is able to recognize the fault FA1, which is consistent with the judgments made by the other of methods except Dempster combination rule and
proves the effectiveness of the proposed method in managing conflicting information provided. Besides, the proposed method owns the biggest value of belief among the judgments on FA1 produced by different methods (0.9265), which is well illustrated in Figure 8 and 9. The reason for
the superior results of combination is because of effectiveness of Z-number in determining contribution proportion, accuracy of OWA operator in distinguishing useful information and reallocation of the mass of evidences based on Yager’s rule of combination. In one word, the proposed method possesses a superior and advantages compared with other existing methods.

\section{Conclusion}
The method proposed in this paper provides a completely new vision in combining conflicting or normal evidences by mainly taking Z-numbers and the concept of pignistic transformation into consideration. From the results obtained in numerical examples and application on real applications, it can be easily concluded that the proposed method retains high accuracy
and robustness in handling complex situations expressed by different evidences. All in all, the proposed offers a distinct solution in combining intricate evidences and it can be utilized to manage problems which can not be appropriately overcame by traditional or modern methods.

%\subsubsection*{Acknowledgments}
%The work is partially supported by the National Natural Science Foundation of China (Grant No. 61973332), and by the China Scholarship Council (202206070008).

\bibliographystyle{elsarticle-num}
\bibliography{cite}
\end{document}